\pdfoutput=1

\documentclass[11pt]{article}

\usepackage{authblk}
\usepackage{microtype}
\usepackage{graphicx}
\usepackage{subfigure}
\usepackage{amsmath}
\usepackage{amsthm}
\usepackage{amsfonts}
\usepackage{stfloats}
\usepackage{multirow}
\usepackage{float}
\usepackage{enumitem}
\usepackage{booktabs} 
\usepackage{tablefootnote}

\usepackage{hyperref}

\usepackage{arxiv}

\usepackage{times}
\usepackage{latexsym}

\usepackage[T1]{fontenc}

\usepackage[utf8]{inputenc}

\usepackage{microtype}

\usepackage{xcolor}
\usepackage{pifont}
\usepackage{amssymb}

\usepackage{afterpage}

\newcommand{\xhdr}[1]{\noindent{\bfseries #1}.}

\newcommand{\cut}[1]{}

\newcommand{\nop}[1]{}
\newcommand{\abr}[1]{\textsc{#1}}
\newcommand{\BERTRL}{\abr{BertRL}}

\title{Inductive Relation Prediction by BERT}

\author{\textbf{Hanwen Zha}}
\author{\textbf{Zhiyu Chen}}
\author{\textbf{Xifeng Yan}}
\affil{University of California, Santa Barbara \authorcr \{hwzha, zhiyuchen, xyan\}@cs.ucsb.edu}

\begin{document}
\maketitle{}
\begin{abstract}
Relation prediction in knowledge graphs is dominated by embedding based methods which mainly focus on the transductive setting.  Unfortunately, they are not able to handle inductive learning where unseen entities and relations are present  and cannot take advantage of prior knowledge.  Furthermore, their inference process is not easily explainable.  In this work, we propose an all-in-one solution, called \BERTRL\ (BERT-based Relational Learning), which leverages pre-trained language model and fine-tunes it by taking relation instances and their possible reasoning paths as training samples. \BERTRL outperforms the SOTAs in 15 out of 18 cases in both inductive and transductive settings. Meanwhile, it demonstrates strong generalization capability in few-shot learning and is explainable \footnote{https://github.com/zhw12/BERTRL}.




\end{abstract}


\section{Introduction} \label{sec:intro}

Knowledge graphs (KGs) are essential in a wide range of tasks such as question answering and recommendation systems \cite{ji2020survey}.\nop{ A vast number of facts in the world can be represented as entities and relations between them using the form of triples.}  As many knowledge graphs are substantially incomplete in practice, knowledge graph completion (KGC) becomes a must in many applications \cite{nickel2016review}.

Embedding-based methods such as TransE \cite{transE}, Complex \cite{complex}, ConvE \cite{dettmers2018conve}, RotatE \cite{sun2018rotate} and TuckER \cite{balazevic2019tucker},  achieve the state-of-the-art performance on a few KGC benchmarks.  However, the drawbacks of these approaches are obvious as they are limited to the \textit{transductive} setting where entities and relations need to be seen at training time.  In reality, new entities and relations emerge over time (\textit{inductive} setting).  The cost of retraining may be too high for dynamically populated knowledge graphs. In addition to the inductive setting, explainability, few-shot learning and transfer learning cannot be easily solved by these specialized embedding methods. 


Logical induction methods partially meet the aforementioned need by seeking probabilistic subgraph patterns (GRAIL \cite{grail}), logical rules (AMIE \cite{amie}, RuleN \cite{ruleN}) or their differentiable counterparts (NEURAL-LP \cite{neural-lp}, DRUM \cite{drum}).  The following shows a logical rule which is explainable, can be generalized, and can handle unseen entities,  
\begin{multline} \label{example_rule}
   (x, \textit{president\_of}, y) \wedge (z, \textit{capital\_of}, y)\\
    \rightarrow (x, \textit{work\_at}, z).
\end{multline}
These logical rules introduce inductive ability for predicting missing links in KG. For example, once the rule in \eqref{example_rule} is learned, the model can generalize to other \textit{president}, \textit{capital} and \textit{country}. 


Despite the compelling advantage of the existing logical induction methods, their inductive learning power is limited as it only exploits the structural information while ignoring the textual information associated with entities and relations, and furthermore, prior knowledge carried in these texts. This weakens the model's usability when only small knowledge graphs are available -- a typical few-shot setting. Moreover, none of them can handle unseen but relevant relations in KG completion.



In this work, we propose an all-in-one solution, called \BERTRL\ (BERT-based Relational Learning), a model that combines rule-based reasoning with textual information and prior knowledge by leveraging pre-trained language model, BERT \cite{bert}.  In \BERTRL, we linearize the local subgraph around entities in a target relation 
$(h, r, t)$ 
into paths $p: (h, r_0, e_1), (e_1, r_1, e_2), \ldots, (e_n, r_n, t)$, input $(h, r, t): p$ to BERT, and then fine-tune. \BERTRL\ is different from  KG-BERT \cite{kgbert} where only relation instance $(h, r, t)$ is fed to BERT.    While this difference looks small, it actually lets \BERTRL\ reason explicitly via paths connecting two entities.  KG-BERT's prediction is mainly based on the representation of entities and relations:  Knowledge graph is memorized inside BERT and reasoning is implicit.  In \BERTRL, knowledge is dynamically retrieved from the knowledge graph during inference: Reasoning is conducted explicitly, which enables \BERTRL\ to achieve explainability and much higher accuracy.  Table \ref{table.comparison} illustrates the difference among these approaches.    

Our approach naturally generalizes to unseen entities.  It also has the potential to handle some unseen relations.  Empirical experiments on inductive knowledge graph completion benchmarks demonstrate the superior performance of \BERTRL\ in comparison with state-of-the-art baselines: It achieves an absolute increase of 6.3\% and 6.5\% in Hits@1 and MRR on average. In a few-shot learning scenario, it can even achieve a maximum of 32.7\% and 27.8\% absolute Hits@1 and MRR improvement.  

In the transductive setting, \BERTRL\  performs competitively with the state-of-the-art embedding methods and surpasses the inductive learning counterparts. In few-shot learning (partially transductive), \BERTRL\ again introduces a larger margin over the baselines. 

Finally, we analyze how \BERTRL\ performs in unseen relation prediction, its explainability, its training and inference time, and conduct an ablation study on a few design choices. 






\begin{table*}
\small 
    \begin{center}
    \setlength{\tabcolsep}{1mm}{
        \begin{tabular}{l | c | c | c | c | c | c}
        \hline
\multicolumn{1}{l|}{\multirow{3}{*}{\textbf{Method}}} & \multicolumn{1}{c|}{\multirow{3}{*}{\textbf{\begin{tabular}[c]{@{}c@{}}Transductive\\ Setting\end{tabular}}}} & \multicolumn{3}{c|}{\textbf{Inductive Setting}} &
\multicolumn{1}{c|}{\multirow{3}{*}{\textbf{\begin{tabular}[c]{@{}c@{}}Prior\\ Knowledge\end{tabular}}}} &
\multicolumn{1}{c}{\multirow{3}{*}{\textbf{Explainable}}}\\ \cline{3-5}
\multicolumn{1}{l|}{} & \multicolumn{1}{c|}{} & \multicolumn{1}{l|}{\textbf{Unseen}} & \multicolumn{1}{l|}{\textbf{Unseen}} & \multicolumn{1}{l|}{\textbf{Reasoning}} & \multicolumn{1}{c|}{} \\
\multicolumn{1}{l|}{} & \multicolumn{1}{c|}{} & \multicolumn{1}{l|}{\textbf{Entities}} & \multicolumn{1}{l|}{\textbf{Relations}} & \multicolumn{1}{l|}{\textbf{with context}} & \multicolumn{1}{c|}{} \\ \hline
\hline

        TuckER & $\checkmark$ & $\times$ & $\times$ & $\times$ & $\times$ & $\times$\\
        RuleN & $\checkmark$ & $\checkmark$ & $\times$ & $\checkmark$ & $\times$ & $\checkmark$\\
        GRAIL & $\checkmark$ & $\checkmark$ & $\times$ & $\checkmark$ & $\times$ & $\times$\\
        KG-BERT & $\checkmark$ & $\checkmark$ & $\checkmark$ & $\times$ & $\checkmark$ & $\times$\\
        \hline
        BERTRL (ours) & $\checkmark$ & $\checkmark$ & $\checkmark$ & $\checkmark$ & $\checkmark$ & $\checkmark$\\

          \hline
           
        \end{tabular}}
    \end{center}
    \caption{Comparison of \BERTRL\ with other relation prediction algorithms on their capability of handling the  transductive setting, unseen entities in the inductive setting, their potential of dealing with unseen relations, usage of prior knowledge, the explainability of their inference process, and whether they can reason with the context of entities in the knowledge graph explicitly.  We take TuckER as a representative of embedding-based methods.}
    \label{table.comparison}
\end{table*}

\section{Proposed Approach}

\begin{figure*}
    \begin{minipage}[b]{\linewidth}
  \centering
  \centerline{\includegraphics[width=\linewidth]{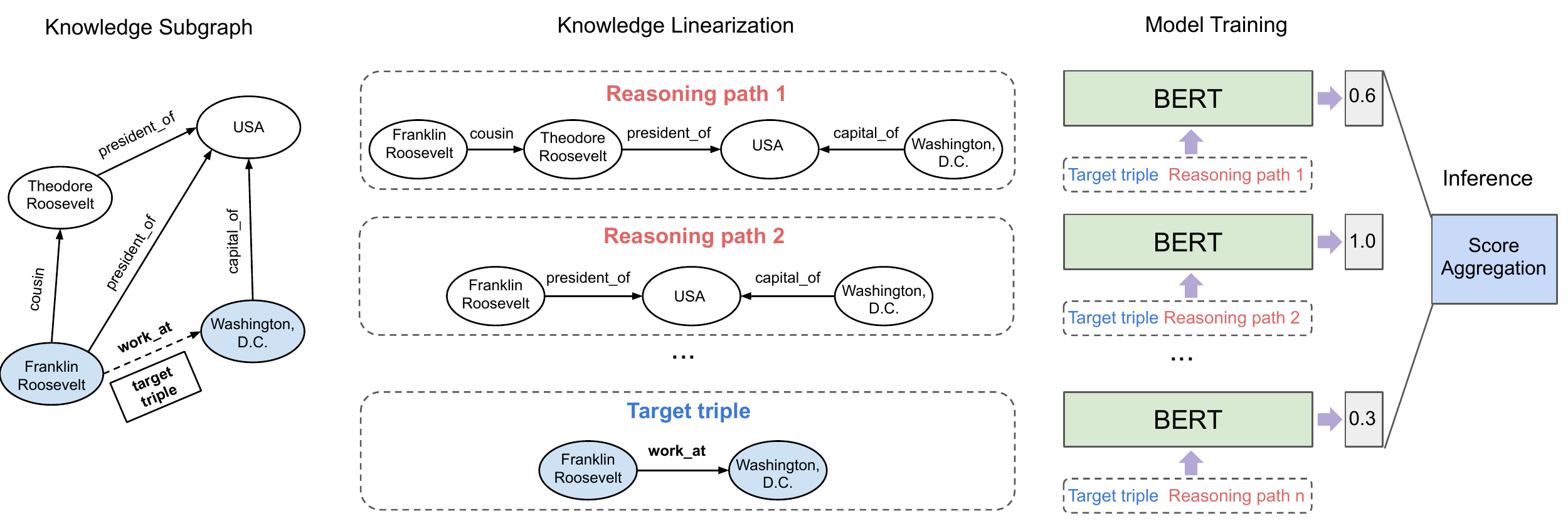}}
\end{minipage}
\caption{The \BERTRL\ pipeline. }

\label{fig:modoverview}
\vspace{-10pt}
\end{figure*}

\xhdr{Problem Formulation}
Knowledge graph consists of a set of triples $\{(h_i, r_i, t_i)\}$ with head, tail entities $h_i, t_i \in \mathcal{E}$ (the set of entities) and relation $r_i \in \mathcal{R}$ (the set of relations). Given an incomplete knowledge graph $G$, the relation prediction task is to score the probability that an unseen relational triple $(h, r, t)$  is true, where $h$ and $t$ denote head and tail entities and $r$ refers to a relation.  $(h, r, t)$ is also called target relational triple. 

Our model scores a relational triple in two steps: (Step 1) Extracting and linearizing the knowledge $G(h, t)$ surrounding entities $h$ and $t$ in $G$; (Step 2) Scoring the triple with $G(h, t)$ by fine-tuning the pre-trained language model BERT.  \nop{Figure \ref{fig:modoverview}b.}

\nop{The key idea of our method is to predict relation by combining structural reasoning and textual pattern association.  Our method leverages textual representation of both target triple and reasoning paths between entities as input for pre-trained language model BERT \cite{bert}.
With the advantage of the contextual representation of BERT, our method naturally generalizes to the unseen entity by taking words or subwords as input. This bypasses the unseen entity embedding problem of traditional embedding based method used in transductive setting.
}





\subsection{Model Details}\label{sec:model}


\xhdr{Step 1: Knowledge Linearization}
The knowledge $G(h, t)$ surrounding entities $h$ and $t$ in a knowledge graph $G$ provides important clues for predicting missing links between $h$ and $t$. $G(h, t)$ could be exploited in various ways: It could be any subgraph around $h$ and $t$ and even not necessarily be connected.  However, the different choices of $G(h, t)$ will affect the model complexity and its explainability. RuleN \cite{ruleN} uses all the paths connecting $h$ and $t$ up to $k$ length. Grail \cite{grail} uses a subgraph that merges all of these paths, aiming to leverage structural information.  In order to use pre-trained language models like BERT, we need to linearize $G(h,t)$ as $\ell(G(h,t))$ and concatenate it with $(h, r, t)$ as valid input to BERT, 
\begin{equation} 
(h, r, t) : \ell(G(h,t)) . 
\end{equation}
Our intuition is that BERT shall have the capability of learning signals in $G(h,t)$ that could be correlated with $(h, r, t)$, and BERT shall be able to handle noisy and erroneous inputs. 


\nop{Subgraph is another alternative of local structural context. The enclosing subgraph, shown effectiveness in GRAIL \cite{grail}, contains nodes located in $k$-hop neighborhood of both head entity $h$ and tail entity $t$. 

Though subgraph preserves the structural context expressively, it can not be directly inputted to a pre-trained language model trained with sequential text.}

\medskip
\xhdr{Subgraph} One straightforward linearization of a subgraph would be concatenating text of its edges one by one separated by a delimiter such as a semicolon.
This formalism has two major issues. First, local subgraphs could be very large: The size grows exponentially with respect to their diameters. Hence concatenated edges may not fit into the available BERT models. Second, the subgraph edges are unordered, which might incur additional cost for BERT to learn orders and produce correct scoring.  We will show experiment of subgraph-based linearization design in Section \ref{sec:abalation}. 



\xhdr{Paths} Another linearization method is collecting all of the paths up to length $k$ connecting $h$ and $t$. We call them \textit{reasoning paths}. 
Each reasoning path between $h$ and $t$ consists of a sequence of triples  $h \to t: (h, r_0, e_1), (e_1, r_1, e_2), ... ,(e_n, r_n, t)$.  

There are two ways of leveraging reasoning paths: 
One called \textit{combined paths}, puts all the paths together as one input to BERT, thus allowing the interaction across different path units. The other called \textit{individual paths}, takes each path as a separate input to BERT. Each reasoning path induces the target triple individually with a certain confidence score, and the final result is an aggregation of individual scores.  In practice, the first method generates one sample concatenating all paths, while the second one separates each path into individual training samples. 

Intuitively, the \textit{combined paths} representation is more expressive as it could consider all the paths together and should perform better.  The \textit{individual paths} representation might generate many false associations as most of the paths are irrelevant to the target triple.  Surprisingly, we found BERT is robust to those false associations taken in the training stage and is able to pick up true ones.
We suspect that the individual paths representation has simpler training samples and likely most relation predictions can be achieved by one path in the existing KGC benchmarks. 

Our final design takes the individual paths representation.  The performance of different designs is presented in Section \ref{sec:abalation}. 

In order to better leverage the knowledge learned in a pre-trained language model, we adopt natural question patterns \cite{schick2020s}.  Take Figure \ref{fig:modoverview} as an example.  It could be  ``[CLS] Question: Franklin Roosevelt work at what ? Is the correct answer Washington D.C. ? [SEP] Context: Franklin Roosevelt president of USA; Washington D.C. capital of USA;''  Each individual path will form a training/inference instance. 







\medskip
\xhdr{Step 2: BERT Scoring} 
In \BERTRL, since we take individual paths as a linearization approach, each pair of triple and reasoning path is scored individually. For each target triple, one or a few reasoning paths would indicate the truth of the triple. This forms a multi-instance learning problem \cite{carbonneau2018multiple}, where predictions need to be aggregated for a bag of instances. 
We take a simplified realization - training individually and applying maximum aggregation of bag scoring at inference time. 

\BERTRL\ uses a linear layer on top of [CLS] to score the triple's correctness, which can be regarded as a binary classification problem. It models the probability of label $y$ ($y \in\{0, 1 \}$) given the text of triple  $(h, r, t)$ and the text of reasoning path  $h \to t$, 
\begin{equation}
    p(y|h, r, t, h \to t).
\end{equation} 
At inference time, the final score of a target triple $(h, r, t)$ is the maximum of the positive class scores over all of its reasoning paths:
\begin{equation}
    score(h, r, t) = \max_{\rho} p(y=1|h, r, t,h\stackrel{\rho}{\to}t). 
\end{equation}
\nop{where $\mathcal{P}$ denotes a set of  reasoning paths between $h$ and $t$.}
The path corresponding to the maximum score can be used to explain how the prediction is derived.  We leave a more sophisticated aggregation function for future study.  

\subsection{Training Regime} \label{sec:training_regime}
In order to train \BERTRL, both positive and negative examples are needed. We follow the standard practice to view existing triples in KG as positive. Then, for each positive triple, we do \textit{negative sampling} to sample $m$ triples corrupting its head or tail. Specifically, we randomly sample entities from common $k$-hop neighbors of head and tail entities, and make sure negative triples are not in KG.
We do not include empty reasoning path examples in training, and always give a minimum confidence score for empty path in inference. 



When constructing reasoning paths for a triple, we hide the triple in KG and find other paths to simulate missing link prediction. As the maximum length of the reasoning paths increases, the number of paths may grow exponentially. Many paths are spurious and not truly useful for inducing the triple. We do \textit{path sampling} at training time to get at most $n$ paths between target entities and take shorter paths first.

Finally we use cross entropy loss to train our model:
\begin{equation}
\mathcal{L} = -\sum_{\tau}{(y_{\tau}\log p_{\tau} + (1 - y_{\tau})\log(1-p_{\tau}))},
\end{equation}
where $y_{\tau} \in \{0,1\}$ indicates negative or positive label, and $\tau \in \mathbb{D}^+ \cup \mathbb{D}^-$. The negative triple set $\mathbb{D}^-$ is generated by previously mentioned method that corrupts head $h$ or tail entity $t$ in a positive triple $(h,r,t) \in \mathbb{D}^+$ with a sampled entity $h'$ or $t'$, i.e., 
\begin{equation}
\begin{aligned}
    \mathbb{D}^- = \{(h',r, t) \notin \mathbb{D}^+  \cup (h,r,t') \notin \mathbb{D}^+\}.
\end{aligned}
\end{equation}





\section{Experiments}

We evaluate our method on three benchmark datasets: WN18RR \cite{dettmers2018conve}, FB15k-237 \cite{toutanova2015representing}, and NELL-995 \cite{xiong2017deeppath}, using their inductive and transductive subsets introduced by \cite{grail} \footnote{https://github.com/kkteru/grail}. WN18RR is a subset of WordNet, a KG contains lexical relations between words. FB15k-237 is a subset of Freebase, a large KG of real-world facts. 
NELL-995 is a dataset constructed from high-confidence facts of NELL, a system constantly extracting facts from the web. 
The statistics of these datasets are given in Table \ref{table:data-stats}; the details of the variants will be given later.

\begin{table}[tp]
\small
    \caption{Statistics of the three datasets and their variants.}
    \centering
    \begin{tabular}{@{}lllll@{}}
        \toprule
 & split & \#relations & \#nodes & \#links \\
 \midrule
\multirow{3}{*}{WN18RR} & train & 9 & 2,746 & 6,670 \\
 & ind-test & 8 & 922 & 1,991 \\
 & train-1000 & 9 & 1,362 & 1,001 \\
 & train-2000 & 9 & 1,970 & 2,002 \\
\midrule
\multirow{2}{*}{FB15k-237} & train & 180 & 1,594 & 5,223 \\
 & ind-test & 142 & 1,093 & 2,404 \\
 & train-1000 & 180 & 923 & 1,027 \\
  & train-2000 & 180 & 1,280 & 2,008 \\
  & train-rel50 & 50 & 1,310 & 3,283 \\
   & train-rel100 & 100 & 1,499 & 3,895 \\
\midrule
\multirow{2}{*}{NELL-995} & train & 88 & 2,564 & 10,063 \\
 & ind-test & 79 & 2,086 & 5,521 \\
 & train-1000 & 88 & 893 & 1,020 \\
 & train-2000 & 88 & 1,346 & 2,011 \\
 \bottomrule
    \end{tabular}
    \label{table:data-stats}
\end{table}



Through experiments, we would like to answer the following questions about \BERTRL: (1) How does it generalize to relation prediction with unseen entities in the inductive setting? (2) How does it perform in the traditional transductive setting? (3) Does it work well in few-shot learning? (4) Does it have the potential to generalize to unseen relations? (5) How its reasoning path explains prediction? (6) What is the training and inference time? (7) How important is the knowledge linearization design?


    
    
    
    
    

\begin{table*}[h]
\small
    \caption{Inductive results (Hits@1)}
    \centering
    \begin{tabular}{@{}llllllllll@{}}
        \toprule
         & \multicolumn{3}{c}{WN18RR} & \multicolumn{3}{c}{FB15k-237} & \multicolumn{3}{c}{NELL-995}\\
         
         \cmidrule(lr){2-4} \cmidrule(lr){5-7} \cmidrule(lr){8-10} \\
        \addlinespace[-12pt]
        & 1,000 & 2,000 & 6,678 (full)  & 1,000 & 2,000 & 5,223 (full)  & 1,000 & 2,000 & 10,063 (full) \\
        \midrule
RuleN & 0.649 & 0.737 & 0.745 & 0.207 & 0.344 & 0.415 & 0.282 & 0.418 & 0.638 \\
GRAIL & 0.516 & \textbf{0.769} & \textbf{0.769} & 0.273 & 0.351 & 0.390 & 0.295 & 0.298 & 0.554 \\
KG-BERT & 0.364 & 0.404 & 0.436 & 0.288 & 0.317 & 0.341 & 0.236 & 0.236 & 0.244 \\
BERTRL & \textbf{0.713} & 0.731 & 0.755 & \textbf{0.441} & \textbf{0.493} & \textbf{0.541} & \textbf{0.622} & \textbf{0.628} & \textbf{0.715} \\
        \bottomrule
    \end{tabular}
    \label{tab:ind_results_hits}
\end{table*}

\begin{table*}[h]
\small
    \caption{Inductive results (MRR)}
    \centering
    \begin{tabular}{@{}llllllllll@{}}
        \toprule
         & \multicolumn{3}{c}{WN18RR} & \multicolumn{3}{c}{FB15k-237} & \multicolumn{3}{c}{NELL-995}\\
         
         \cmidrule(lr){2-4} \cmidrule(lr){5-7} \cmidrule(lr){8-10} \\
        \addlinespace[-12pt]
        & 1,000 & 2,000 & 6,678 (full)  & 1,000 & 2,000 & 5,223 (full)  & 1,000 & 2,000 & 10,063 (full) \\
        \midrule
RuleN & 0.681 & 0.773 & 0.780 & 0.236 & 0.383 & 0.462 & 0.334 & 0.495 & 0.710 \\
GRAIL & 0.652 & \textbf{0.799} & \textbf{0.799} & 0.380 & 0.432 & 0.469 & 0.458 & 0.462 & 0.675 \\
KG-BERT & 0.471 & 0.525 & 0.547 & 0.431 & 0.460 & 0.500 & 0.406 & 0.406 & 0.419 \\
BERTRL & \textbf{0.765} & 0.777 & 0.792 & \textbf{0.526} & \textbf{0.565} & \textbf{0.605} & \textbf{0.736} & \textbf{0.744} & \textbf{0.808} \\
        \bottomrule
    \end{tabular}
    \label{tab:ind_results_mrr}
\end{table*}

\medskip
\xhdr{Baselines and Implementation Details}
We compare \BERTRL\ with the state-of-the-art inductive relation prediction methods GRAIL \cite{grail} and RuleN \cite{ruleN}. GRAIL uses graph neural network to reason over local subgraph structures. RuleN explicitly derives path-based rules and shows high precision. We use the public implementation provided by the authors and adopt the best hyper-parameter settings in their work.  Differentiable logical rule learning methods like NeurLP \cite{neural-lp} and DRUM \cite{drum} are not included, as their performance is not as good as GRAIL and RuleN \cite{grail}. For the transductive setting, we pick one of the state-of-the-art embedding methods, TuckER \cite{balazevic2019tucker} and path-based method MINERVA \cite{minerva}, as representatives for evaluation. For TuckER, we use implementation in LibKGE \cite{broscheit2020libkge} with the provided best configuration in the library. For MINERVA, we use the official implementation and best configuration provided by authors. 

We also compare against a BERT-based KGC method KG-BERT \cite{kgbert}, where only relation triple $(h, r, t)$ is fed to BERT. This is a special case of \BERTRL\ with an empty reasoning path. In our experiments, we do not feed additional description other than entity and relation names as \cite{kgbert} did.  We aim to give all the methods the same input.  In practice, both can be extended to accept additional information as this is what  BERT is designed for.

Both \BERTRL\ and KG-BERT were implemented in Pytorch using Huggingface Transformers library \cite{huggingface}. We employ BERT base model (cased) with 12 layers and 110M parameters and run experiments with a GTX 1080 Ti GPU with 12GB RAM. We use a batch size of 32 and fine-tune models for 2 epochs using the Adam optimizer.  The best learning rate 5e-5 is set for \BERTRL\ and 2e-5 for KG-BERT, selected from 2e-5 to 5e-5 based on validation set performance. We sample 10 negative triples in negative sampling, and 3 reasoning paths in path sampling. We set other hyperparameters as their default values in the package.



\medskip
\xhdr{Evaluation Task} Following GRAIL \cite{grail}, our default evaluation task is to rank each test triple among 50 other negative candidates. The negative triples are not in KG and generated by randomly replacing head (or tail) entity of each test triple. The sampling is going to speed up the evaluation process. The performance will be lower if the ranking is done among the full entity set.


\medskip
\xhdr{Metrics} We evaluate the models on Hits@1 and Mean Reciprocal Rank (MRR). Hits@1 measures the percentage of cases in which positive triple appears as the top 1 ranked triple, while MRR takes the average of the reciprocal rank for positive triples. 

\subsection{Inductive Relation Prediction}
We first evaluate the model's ability to generalize to unseen entities.  In a fully inductive setting, the entities seen in training and testing are completely disjoint.  For all the methods, we extract paths from the target head entity to the  tail entity with length up to 3 or the subgraph containing these paths.
\nop{In practice, fast-evolving knowledge graph in many cases involve small training data especially for new relations. The fewer entities seen in the training, the better the model's learning capability can be characterized -- a typical few-shot learning scenario. }

\medskip
\xhdr{Datasets} We conduct our experiment using the inductive subsets of WN18RR, FB15k-237, and NELL-995 introduced by \cite{grail}. Each subset consists of a pair of graphs \textit{train-graph} and \textit{ind-test-graph}. The former is used for training, and the latter provides an incomplete graph for relation prediction.  \textit{train-graph} contains all the relations present in \textit{ind-test-graph}. However, their entity sets do not overlap.  In GRAIL, WN18RR, FB15k-237, and NELL-995 each induces four random inductive subsets (v1, v2, v3 and v4). We pick one subset for each (WN18RR v1, FB15k-237 v1 and NELL-995 v2).
For each inductive dataset, we did stratified sampling on \textit{train-graph} to create few-shot variants. 
The links are down-sampled to a number around 1,000 and 2,000, while keeping an unchanged proportion of triples for each relation. The few-shot training graph \textit{train-1000} and \textit{train-2000} contain all relations in its full setting, thus covering the relations in \textit{test-graph} as well.  The statistics of these variants are shown in Table \ref{table:data-stats}.

\begin{table*}[h]
\small
    \caption{Transductive results (Hits@1)}
    \centering
    \begin{tabular}{@{}llllllllll@{}}
        \toprule
 & \multicolumn{3}{c}{\textbf{Transductive}} & \multicolumn{6}{c}{\textbf{Transductive (Few-shot)}
 } \\ 
 \cmidrule(lr){2-4}   \cmidrule(lr){5-10} \\
\addlinespace[-12pt]
         &WN18RR & FB15k-237 & NELL-995 & \multicolumn{2}{c}{WN18RR} & \multicolumn{2}{c}{FB15k-237} & \multicolumn{2}{c}{NELL-995}\\
  \cmidrule(lr){2-4} \cmidrule(lr){5-6} \cmidrule(lr){7-8} \cmidrule(lr){9-10} \\
        \addlinespace[-12pt]
        & 6,670 & 5,223 & 10,063 & 1,000 & 2,000 & 1,000  & 2,000 & 1,000 & 2,000 \\
        \midrule
RuleN & 0.646 & 0.603 & 0.636 & 0.548 & 0.605 & 0.374 & 0.508 & 0.365 & 0.501 \\
GRAIL & 0.644 & 0.494 & 0.615 & 0.489 & 0.633 & 0.267 & 0.352 & 0.198 & 0.342 \\
MINERVA & 0.632 & 0.534 & 0.553 & 0.106 & 0.248 & 0.170 & 0.324 & 0.152 & 0.284 \\
TuckER & 0.600 & 0.615 & \textbf{0.729} & 0.230 & 0.415 & 0.407 & 0.529 & 0.392 & 0.520 \\
BERTRL & \textbf{0.655} & \textbf{0.620} & 0.686 & \textbf{0.621} & \textbf{0.637} & \textbf{0.517} & \textbf{0.583} & \textbf{0.526} & \textbf{0.582} \\
        \bottomrule
    \end{tabular}
    \label{tab:trans_results_hits}
\end{table*}


\begin{table*}[h]
\small
    \caption{Transductive results (MRR)}
    \centering
    \begin{tabular}{@{}llllllllll@{}}
        \toprule
 & \multicolumn{3}{c}{\textbf{Transductive}} & \multicolumn{6}{c}{\textbf{Transductive (Few-shot)
 }} \\ 
 \cmidrule(lr){2-4}   \cmidrule(lr){5-10} \\
\addlinespace[-12pt]
         &WN18RR & FB15k-237 & NELL-995 & \multicolumn{2}{c}{WN18RR} & \multicolumn{2}{c}{FB15k-237} & \multicolumn{2}{c}{NELL-995}\\
  \cmidrule(lr){2-4} \cmidrule(lr){5-6} \cmidrule(lr){7-8} \cmidrule(lr){9-10} \\
        \addlinespace[-12pt]
        & 6,670 & 5,223 & 10,063  & 1,000 & 2,000 & 1,000  & 2,000 & 1,000 & 2,000 \\
        \midrule
RuleN & 0.669 & 0.674 & 0.736 & 0.567 & 0.625 & 0.434 & 0.577 & 0.453 & 0.609 \\
GRAIL & 0.676 & 0.597 & 0.727 & 0.588 & 0.673 & 0.375 & 0.453 & 0.292 & 0.436 \\
MINERVA & 0.656 & 0.572 & 0.592 & 0.125 & 0.268 & 0.198 & 0.364 & 0.182 & 0.322 \\
TuckER & 0.646 & 0.682 & \textbf{0.800} & 0.258 & 0.448 & 0.457 & 0.601 & 0.436 & 0.577 \\
BERTRL & \textbf{0.683} & \textbf{0.695} & 0.781 & \textbf{0.662} & \textbf{0.673} & \textbf{0.618} & \textbf{0.667} & \textbf{0.648} & \textbf{0.693} \\
        \bottomrule
    \end{tabular}
     \label{tab:trans_results_mrr}
\end{table*}

\medskip
\xhdr{Results} \BERTRL\ significantly outperforms the baselines in most settings as shown in Tables \ref{tab:ind_results_hits} and \ref{tab:ind_results_mrr}, particularly by around 10 absolute Hits@1 and MRR points in FB15k-237 and NELL-995. These two KGs have more relations and are  associated with open-world knowledge (learned by BERT) compared with WN18RR. Methods like GRAIL and RuleN are not able to incorporate such prior knowledge.

In the few-shot setting, \BERTRL\ stays robust and outperforms the baselines by an even larger margin. When more links are dropped in training graph, \BERTRL\ achieves more performance gain over the baselines. \BERTRL\ enjoys all sources of knowledge: structural (reasoning paths), textual (embedding), and prior knowledge (pre-trained language model). They all play an important role in knowledge graph completion. 

In both settings, \BERTRL\ performs better than KG-BERT, the version without reasoning paths inputted.  It shows that incorporating paths allows pre-trained language models to gain explicit reasoning capability.  On the other hand, with the triple information alone, KG-BERT is able to make a certain amount of correct inferences, suggesting that prior knowledge stored in  pre-trained language models can be leveraged to do knowledge graph completion as manifested in \cite{kgbert}.  \BERTRL\ combines explicit reasoning capability, prior knowledge, and language understanding all together in one model and has significant advantages.

\subsection{Transductive Relation Prediction}
\BERTRL\ can also be applied in the transductive setting and be compared with the baselines. 

\medskip
\xhdr{Datasets}
To evaluate the transductive performance, we train these models on \textit{train-graph} introduced in the inductive setting and test on links with the same set of entities. We use a list of test triples with 10\% size of \textit{train-graph}. In a few-shot setting, we reuse the few-shot \textit{train-graph} used in the inductive setting and tested on the aforementioned test links. At testing time, full \textit{train-graph} is used to collect knowledge around target entities (otherwise, the setting will be close to the inductive one).  The few-shot setting makes datasets partially transductive, as some entities become unseen when links are dropped randomly. For TuckER and MINERVA, we assign a minimum score for both positive and negative triples containing unseen entities.


\medskip
\xhdr{Results} Tables \ref{tab:trans_results_hits} and \ref{tab:trans_results_mrr} show that \BERTRL\ outperforms the baselines in most of full and few-shot settings.  It performs competitively with TuckER in the full setting and surpasses RuleN and GRAIL.  It implies that \BERTRL's strong performance is not limited to  inductive learning.  In the few-shot setting, \textit{train-graph} becomes sparse and unseen entities appear in testing.  \BERTRL\ again largely outperforms all the methods, which once more demonstrates the advantage of  simultaneously exploiting all knowledge sources.



\nop{From Table [], we observe than the performance gain is somewhere between the fully transductive setting and fully inductive setting. This again shows that our model's consistent strong performance in various kinds of few-shot scenario.}

\begin{table}[h]
\small
    \caption{Unseen relation prediction results (Hits@1)}
    \centering
    \begin{tabular}{@{}lll@{}}
        \toprule
         &  50 relations & 100 relations \\
	    \midrule
    KG-BERT & 0.266 & 0.450  \\
    BERTRL & 0.485 & 0.500 \\
        \bottomrule
    \end{tabular}
     \label{tab:unseen}
\end{table}

\begin{table*}[h]
\small
\caption{Examples of the best and worst preforming unseen relation prediction of \BERTRL,  trained on a 50 relations subset of FB15k-237.}
    \begin{tabular}{lll}
    \toprule
    Unseen relation & Hits@1 & Similar seen relation \\
    \midrule
    /film/film\_format & 1.000 & /film/genre, /film/language \\
    /person/spouse\_s./marriage/spouse & 1.000 & /person/spouse\_s./marriage/type\_of\_union \\
    /pro\_athlete/teams./sports\_team\_roster/team & 1.000 & /football\_player/current\_team./sports\_team\_roster/team \\
    \midrule
    \midrule
    /artist/origin & 0.000 & - \\
    /record\_label/artist & 0.100 & - \\
    /ethnicity/languages\_spoken & 0.250 & /person/languages \\
    \bottomrule
    \end{tabular}
    \label{table:unseen_relation_example}
\end{table*}

\subsection{Unseen Relation Prediction}
As \BERTRL\ leverages a pre-trained language model, it has the potential to predict unseen relations in a zero-shot setting, which is not possible for traditional inductive learning methods like RuleN and GRAIL.  In this section, we examine how \BERTRL\ can generalize for unseen relations.

\medskip
\xhdr{Datasets}
We create a down-sampled training dataset from full FB15k-237 \textit{train-graph}, and test  on \textit{ind-test-graph}.  The relations in FB15k-237 have a multi-level hierarchy, e.g.,  people/person/spouse\_s.  Words are shared across different relations, which makes unseen relation generalization possible.
When down-sampling \textit{train-graph}, we sample 50 and 100 relations without replacement weighted by their proportion in \textit{train-graph}, written as \textit{train-rel50} and \textit{train-rel100}. 


\medskip
\xhdr{Results}
Table \ref{tab:unseen} shows Hits@1 results.  It is observed that both KG-BERT and \BERTRL\  make some correct predictions even without seeing the relations in training: The textual information shared among relation names benefits the reasoning of unseen relations. Certainly, both methods take advantage of the knowledge learned by BERT.

Table \ref{table:unseen_relation_example} shows the best and worst performed unseen relation prediction on \textit{train-rel50}. For each unseen relation, we manually identify relevant relations showing in the training set.  These examples show that the best performing relations have some close meaning counterparts seen in training.  In contrast, the worst performing relations are usually distant from relations seen in training. This phenomenon indicates that in the  zero-shot setting, \BERTRL\ generalizes best to unseen but closely relevant relations.  We suspect that knowledge captured by pre-trained language models also helps zero-shot learning. 



\subsection{Explainability}
As stated in Section \ref{sec:intro}, rules like \eqref{example_rule} are explainable to humans. \BERTRL\ achieves certain explainability by leveraging reasoning paths and implicitly memorizes these rules through training.
For a prediction task $(h, r, ?)$,   \BERTRL\ is going to generate many instances for different tail entity $t$ by concatenating triple $(h,r,t)$ with each path $h \to t$. 
Those with the highest scores are chosen as the answer.  We can regard the path chain as the explanation of deriving $(h, r, t)$. We conduct manual case study using FB15k-237 dataset as an example. The texts are simplified.

The following KG completion query \textit{(Chris, acts\_in\_film, ?)} is to find what film the actor Chris acts in. The instance ranked highest by \BERTRL\ consists of target triple \textit{(Chris, acts\_in\_film, Jackie Brown)}, reasoning path \textit{(Chris, nominated\_for\_same\_award\_with, Robert); (Robert, acts\_in\_film, Jackie Brown);} and an assigned score $0.95$. 
It could be naturally explained as follows:  Chris likely acts in film Jackie Brown, since Robert shares the same award nomination with Chris and also acts in Jackie Brown. 

We then examined the percentage of the explanations that do make sense.   We randomly sampled 100 test triples from FB15k-237 and ask human annotators to check their top-1 path chains highly scored by \BERTRL. Human judges found that 84\% of the path chains make sense, indicating strong explainability. 


%

\subsection{Training and Inference Time}
\begin{figure}[t]
\begin{minipage}[b]{0.96\linewidth}
  \centering
  \centerline{\includegraphics[width=\linewidth]{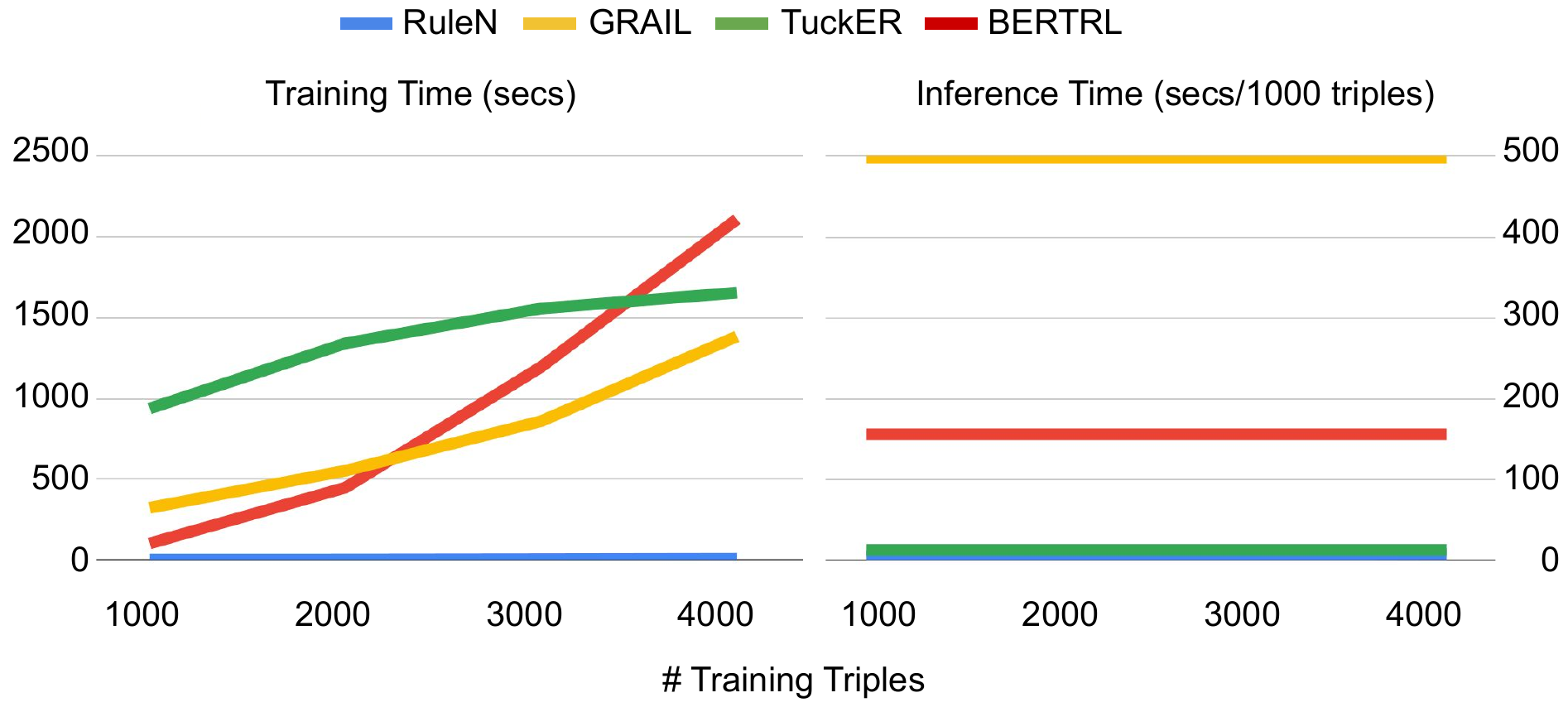}}
\end{minipage}
\caption{Training and inference time with respect to number of training triples. Inference time of TuckER is slightly higher than RuleN and the curves are overlapped.}
\vspace{-10pt}
\label{fig:running_time}
\end{figure}

 We investigate training and inference time, using the transductive setting of FB15k-237 as an example. Figure \ref{fig:running_time} shows the running time of \BERTRL\ compared with other methods using their default packages without further optimization. 
 The running time is highly implementation and device dependent, however, the curves still show a trend and gives a rough scale of it.
 The training time of \BERTRL\ gradually increases as the number of training triples grows. The inference time of \BERTRL\ does not depend on the training data size and is slower than RuleN. Running time is one important factor in practice, and we leave how to speed up \BERTRL\ to future work.
 
 

 


\subsection{Ablation study} \label{sec:abalation}
Table \ref{tab:ablation} shows the effect of different design choices in \BERTRL, mainly knowledge linearization and path sampling. We use the FB15k-237 inductive dataset and its few-shot subset for evaluation.

\medskip
\xhdr {Combined Paths}
As discussed in Section \ref{sec:training_regime}, \textit{combined paths} is one way linearizing structural knowledge.  Although it includes more information in one input, it does not outperform \textit{individual paths}.  This indicates that BERT struggles to learn from complex input when training data is limited, which might be explained by Occam's razor.  
\nop{Splitting reasoning paths into different examples introduces a stronger inductive bias:  reasoning path can affect reasoning individually. This eases learning the association between target relation instance and the reasoning path in \BERTRL.}

\medskip
\xhdr {Subgraph (Edge List)} 
Edge list is the worst performing linearization option.  Linking entities in the input and then recognizing patterns could be more challenging for BERT than reasoning along paths where edges are ordered by their connection. 


\medskip
\xhdr {Path Sampling}
We evaluate the performance of \textit{path sampling} by randomly selecting $n$ paths between entities.  Path sampling could speed up training as the training data becomes small.  The performance is still good even when the number of sampled paths is very small, indicating \BERTRL\ is robust to the size of the training set.  



\begin{table}[t]
\small
    \caption{Ablation study of BERTRL variants (Hits@1)}
    \centering
    \begin{tabular}{@{}llll@{}}
        \toprule
         & 1,000 & 2,000 & full \\
	\midrule

Subgraph (edge list) & 0.361 & 0.398 & 0.463 \\
Combined paths & 0.351 & 0.461 & 0.505 \\
5 sampled individual paths & 0.466 & 0.490 & 0.532 \\
10 sampled individual paths & 0.449 & 0.505 & 0.500 \\
BERTRL (individual paths) & 0.441 & 0.493 & 0.541 \\
        \bottomrule
    \end{tabular}
    \label{tab:ablation}
\end{table}



\section{Related Work}
\xhdr{Transductive Models}
Most existing knowledge graph completion methods are embedding based, such as TransE \cite{transE}, Complex \cite{complex}, ConvE \cite{dettmers2018conve}, RotatE \cite{sun2018rotate} and TuckER \cite{balazevic2019tucker}. These methods learn embedding of entities and relations and construct scoring functions on top of the embedding.  They are naturally transductive and can not be directly applied to or need re-training for the inductive setting where entities are not seen in the training.

Some methods, e.g.,  R-GCN \cite{schlichtkrull2018modeling}, DeepPath \cite{xiong2017deeppath}, MINERVA \cite{minerva} and DIVA \cite{chen2018variational}, learn to aggregate information from local subgraph and paths.  However,  they cannot be directly applied to the inductive setting as entity/node specific embeddings are needed.

\medskip
\xhdr{Inductive Models}
In contrast to the transductive setting, probabilistic rule learning AMIE \cite{amie} and RuleN \cite{ruleN} could apply learned rules to unseen entities.  NeuralLP \cite{neural-lp} and DRUM \cite{drum} learns differentiable rules in an end-to-end manner.  GRAIL \cite{grail} extracts subgraph connecting target entities and learns a general graph neural network to score a prediction. These methods are in nature inductive as they learn entity irrelevant rules or models and conduct reasoning with knowledge graph information only. 

Besides these studies, there are methods learning to generate inductive embedding for unseen nodes. \cite{hamilton2017inductive} and \cite{Bojchevski2017DeepGE} rely on the node features which may not be easily acquired in many KGs.  \cite{lan} and \cite{ookb} generate embedding for unseen nodes by learning to aggregate embedding from neighbors using GNNs. However, those two paradigms require a certain number of known entities and cannot be applied to entirely new graphs.

\medskip
\xhdr{Pre-trained Language Models}
Pre-trained language model is one of the most influential advances in natural language processing, e.g., BERT \cite{bert}, Roberta \cite{liu2019roberta}, and  GPT \cite{radford2019language, brown2020language}.  They are trained unsupervisedly on very large corpus and often achieve great performance after fine-tuning on downstream tasks. Besides, \cite{petroni2019language} introduces LAMA benchmark, and shows that pre-trained language models themselves already capture some factual knowledge even without fine-tuning.

KG-BERT \cite{kgbert} aims to leverage the power of pre-trained language model in knowledge graph completion, where it represents triples as text sequences and uses BERT to learn scoring function for relation prediction.  Though it can be applied in the inductive setting, its prediction is mainly based on the pre-trained representation of entities and relations; it does not learn a general reasoning mechanism like GRAIL and \BERTRL. 

\section{Conclusion}
We proposed \BERTRL, a pre-trained language model based approach for knowledge graph completion. By taking reasoning path and triple as input to a pre-trained language model, \BERTRL\ naturally handles unseen entities and gains the capability of relational reasoning.  In few-shot learning, it outperforms competitive baselines by an even larger margin.  It has the potential to generalize to unseen relations in a zero-shot setting. It not only achieves the state-of-the-art results in inductive learning, but also shown to be effective in transductive learning.  Overall, this work opens a new direction of combining the power of pre-trained language model and logic reasoning.

\bibliography{references}

\begin{thebibliography}{31}
\expandafter\ifx\csname natexlab\endcsname\relax\def\natexlab#1{#1}\fi

\bibitem[{Bala\v{z}evi\'c et~al.(2019)Bala\v{z}evi\'c, Allen, and
  Hospedales}]{balazevic2019tucker}
Ivana Bala\v{z}evi\'c, Carl Allen, and Timothy~M Hospedales. 2019.
\newblock Tucker: Tensor factorization for knowledge graph completion.
\newblock In \emph{Empirical Methods in Natural Language Processing}.

\bibitem[{Bojchevski and G{\"u}nnemann(2018)}]{Bojchevski2017DeepGE}
Aleksandar Bojchevski and Stephan G{\"u}nnemann. 2018.
\newblock Deep gaussian embedding of attributed graphs: Unsupervised inductive
  learning via ranking.
\newblock In \emph{ICLR}.

\bibitem[{Bordes et~al.(2013)Bordes, Usunier, García-Durán, Weston, and
  Yakhnenko}]{transE}
Antoine Bordes, Nicolas Usunier, Alberto García-Durán, Jason Weston, and
  Oksana Yakhnenko. 2013.
\newblock Translating embeddings for modeling multi-relational data.
\newblock In \emph{NIPS}.

\bibitem[{Broscheit et~al.(2020)Broscheit, Ruffinelli, Kochsiek, Betz, and
  Gemulla}]{broscheit2020libkge}
Samuel Broscheit, Daniel Ruffinelli, Adrian Kochsiek, Patrick Betz, and Rainer
  Gemulla. 2020.
\newblock Libkge-a knowledge graph embedding library for reproducible research.
\newblock In \emph{Proceedings of the 2020 Conference on Empirical Methods in
  Natural Language Processing: System Demonstrations}, pages 165--174.

\bibitem[{Brown et~al.(2020)Brown, Mann, Ryder, Subbiah, Kaplan, Dhariwal,
  Neelakantan, Shyam, Sastry, Askell et~al.}]{brown2020language}
Tom~B Brown, Benjamin Mann, Nick Ryder, Melanie Subbiah, Jared Kaplan, Prafulla
  Dhariwal, Arvind Neelakantan, Pranav Shyam, Girish Sastry, Amanda Askell,
  et~al. 2020.
\newblock Language models are few-shot learners.
\newblock \emph{arXiv preprint arXiv:2005.14165}.

\bibitem[{Carbonneau et~al.(2018)Carbonneau, Cheplygina, Granger, and
  Gagnon}]{carbonneau2018multiple}
Marc-Andr{\'e} Carbonneau, Veronika Cheplygina, Eric Granger, and Ghyslain
  Gagnon. 2018.
\newblock Multiple instance learning: A survey of problem characteristics and
  applications.
\newblock \emph{Pattern Recognition}, 77:329--353.

\bibitem[{Chen et~al.(2018)Chen, Xiong, Yan, and Wang}]{chen2018variational}
Wenhu Chen, Wenhan Xiong, Xifeng Yan, and William~Yang Wang. 2018.
\newblock Variational knowledge graph reasoning.
\newblock In \emph{Proceedings of the 2018 Conference of the North American
  Chapter of the Association for Computational Linguistics: Human Language
  Technologies, Volume 1 (Long Papers)}, pages 1823--1832.

\bibitem[{Das et~al.(2018)Das, Dhuliawala, Zaheer, Vilnis, Durugkar,
  Krishnamurthy, Smola, and McCallum}]{minerva}
Rajarshi Das, Shehzaad Dhuliawala, Manzil Zaheer, Luke Vilnis, Ishan Durugkar,
  Akshay Krishnamurthy, Alex Smola, and Andrew McCallum. 2018.
\newblock \href {https://openreview.net/forum?id=Syg-YfWCW} {Go for a walk and
  arrive at the answer: Reasoning over paths in knowledge bases using
  reinforcement learning}.
\newblock In \emph{6th International Conference on Learning Representations,
  {ICLR} 2018, Vancouver, BC, Canada, April 30 - May 3, 2018, Conference Track
  Proceedings}. OpenReview.net.

\bibitem[{Dettmers et~al.(2018)Dettmers, Pasquale, Pontus, and
  Riedel}]{dettmers2018conve}
Tim Dettmers, Minervini Pasquale, Stenetorp Pontus, and Sebastian Riedel. 2018.
\newblock Convolutional 2d knowledge graph embeddings.
\newblock In \emph{AAAI}.

\bibitem[{Devlin et~al.(2019)Devlin, Chang, Lee, and Toutanova}]{bert}
Jacob Devlin, Ming-Wei Chang, Kenton Lee, and Kristina Toutanova. 2019.
\newblock Bert: Pre-training of deep bidirectional transformers for language
  understanding.
\newblock In \emph{NAACL-HLT (1)}.

\bibitem[{Gal\'{a}rraga et~al.(2013)Gal\'{a}rraga, Teflioudi, Hose, and
  Suchanek}]{amie}
Luis~Antonio Gal\'{a}rraga, Christina Teflioudi, Katja Hose, and Fabian
  Suchanek. 2013.
\newblock Amie: Association rule mining under incomplete evidence in
  ontological knowledge bases.
\newblock In \emph{WWW '13}.

\bibitem[{Hamaguchi et~al.(2017)Hamaguchi, Oiwa, Shimbo, and Matsumoto}]{ookb}
Takuo Hamaguchi, Hidekazu Oiwa, Masashi Shimbo, and Yuji Matsumoto. 2017.
\newblock Knowledge transfer for out-of-knowledge-base entities: A graph neural
  network approach.
\newblock In \emph{IJCAI}.

\bibitem[{Hamilton et~al.(2017)Hamilton, Ying, and
  Leskovec}]{hamilton2017inductive}
William~L. Hamilton, Rex Ying, and Jure Leskovec. 2017.
\newblock Inductive representation learning on large graphs.
\newblock In \emph{NIPS}.

\bibitem[{Ji et~al.(2020)Ji, Pan, Cambria, Marttinen, and Yu}]{ji2020survey}
Shaoxiong Ji, Shirui Pan, Erik Cambria, Pekka Marttinen, and Philip~S Yu. 2020.
\newblock A survey on knowledge graphs: Representation, acquisition and
  applications.
\newblock \emph{arXiv preprint arXiv:2002.00388}.

\bibitem[{Liu et~al.(2019)Liu, Ott, Goyal, Du, Joshi, Chen, Levy, Lewis,
  Zettlemoyer, and Stoyanov}]{liu2019roberta}
Yinhan Liu, Myle Ott, Naman Goyal, Jingfei Du, Mandar Joshi, Danqi Chen, Omer
  Levy, Mike Lewis, Luke Zettlemoyer, and Veselin Stoyanov. 2019.
\newblock Roberta: A robustly optimized bert pretraining approach.
\newblock \emph{arXiv preprint arXiv:1907.11692}.

\bibitem[{Meilicke et~al.(2018)Meilicke, Fink, Wang, Ruffinelli, Gemulla, and
  Stuckenschmidt}]{ruleN}
Christian Meilicke, Manuel Fink, Yanjie Wang, Daniel Ruffinelli, Rainer
  Gemulla, and Heiner Stuckenschmidt. 2018.
\newblock Fine-grained evaluation of rule- and embedding-based systems for
  knowledge graph completion.
\newblock In \emph{ISWC}.

\bibitem[{Nickel et~al.(2016)Nickel, Murphy, Tresp, and
  Gabrilovich}]{nickel2016review}
M.~Nickel, K.~Murphy, V.~Tresp, and E.~Gabrilovich. 2016.
\newblock A review of relational machine learning for knowledge graphs.
\newblock \emph{IEEE}.

\bibitem[{Petroni et~al.(2019)Petroni, Rockt{\"a}schel, Riedel, Lewis, Bakhtin,
  Wu, and Miller}]{petroni2019language}
Fabio Petroni, Tim Rockt{\"a}schel, Sebastian Riedel, Patrick Lewis, Anton
  Bakhtin, Yuxiang Wu, and Alexander Miller. 2019.
\newblock Language models as knowledge bases?
\newblock In \emph{Proceedings of the 2019 Conference on Empirical Methods in
  Natural Language Processing and the 9th International Joint Conference on
  Natural Language Processing (EMNLP-IJCNLP)}, pages 2463--2473.

\bibitem[{Radford et~al.(2019)Radford, Wu, Child, Luan, Amodei, and
  Sutskever}]{radford2019language}
Alec Radford, Jeffrey Wu, Rewon Child, David Luan, Dario Amodei, and Ilya
  Sutskever. 2019.
\newblock Language models are unsupervised multitask learners.
\newblock \emph{OpenAI blog}, 1(8):9.

\bibitem[{Sadeghian et~al.(2019)Sadeghian, Armandpour, Ding, and Wang}]{drum}
Ali Sadeghian, Mohammadreza Armandpour, Patrick Ding, and Daisy~Zhe Wang. 2019.
\newblock Drum: End-to-end differentiable rule mining on knowledge graphs.
\newblock In \emph{NeurIPS}.

\bibitem[{Schick and Sch{\"u}tze(2020)}]{schick2020s}
Timo Schick and Hinrich Sch{\"u}tze. 2020.
\newblock It's not just size that matters: Small language models are also
  few-shot learners.
\newblock \emph{arXiv preprint arXiv:2009.07118}.

\bibitem[{Schlichtkrull et~al.(2018)Schlichtkrull, Kipf, Bloem, Van Den~Berg,
  Titov, and Welling}]{schlichtkrull2018modeling}
Michael Schlichtkrull, Thomas~N Kipf, Peter Bloem, Rianne Van Den~Berg, Ivan
  Titov, and Max Welling. 2018.
\newblock Modeling relational data with graph convolutional networks.
\newblock In \emph{European Semantic Web Conference}, pages 593--607. Springer.

\bibitem[{Sun et~al.(2019)Sun, Deng, Nie, and Tang}]{sun2018rotate}
Zhiqing Sun, Zhi-Hong Deng, Jian-Yun Nie, and Jian Tang. 2019.
\newblock Rotate: Knowledge graph embedding by relational rotation in complex
  space.
\newblock In \emph{ICLR}.

\bibitem[{Teru et~al.(2020)Teru, Denis, and Hamilton}]{grail}
Komal~K. Teru, Etienne Denis, and William~L. Hamilton. 2020.
\newblock Inductive relation prediction by subgraph reasoning.
\newblock \emph{arXiv: Learning}.

\bibitem[{Toutanova et~al.(2015)Toutanova, Chen, Pantel, Poon, Choudhury, and
  Gamon}]{toutanova2015representing}
Kristina Toutanova, Danqi Chen, Patrick Pantel, Hoifung Poon, Pallavi
  Choudhury, and Michael Gamon. 2015.
\newblock Representing text for joint embedding of text and knowledge bases.
\newblock In \emph{EMNLP}.

\bibitem[{Trouillon et~al.(2017)Trouillon, Dance, {{\'E}}ric Gaussier, Welbl,
  Riedel, and Bouchard}]{complex}
Th{{\'e}}o Trouillon, Christopher~R. Dance, {{\'E}}ric Gaussier, Johannes
  Welbl, Sebastian Riedel, and Guillaume Bouchard. 2017.
\newblock Knowledge graph completion via complex tensor factorization.
\newblock \emph{JMLR}.

\bibitem[{Wang et~al.(2019)Wang, Han, Li, and Pan}]{lan}
Peifeng Wang, Jialong Han, Chenliang Li, and Rong Pan. 2019.
\newblock Logic attention based neighborhood aggregation for inductive
  knowledge graph embedding.
\newblock In \emph{AAAI}.

\bibitem[{Wolf et~al.(2020)Wolf, Debut, Sanh, Chaumond, Delangue, Moi, Cistac,
  Rault, Louf, Funtowicz, Davison, Shleifer, von Platen, Ma, Jernite, Plu, Xu,
  Scao, Gugger, Drame, Lhoest, and Rush}]{huggingface}
Thomas Wolf, Lysandre Debut, Victor Sanh, Julien Chaumond, Clement Delangue,
  Anthony Moi, Pierric Cistac, Tim Rault, Rémi Louf, Morgan Funtowicz, Joe
  Davison, Sam Shleifer, Patrick von Platen, Clara Ma, Yacine Jernite, Julien
  Plu, Canwen Xu, Teven~Le Scao, Sylvain Gugger, Mariama Drame, Quentin Lhoest,
  and Alexander~M. Rush. 2020.
\newblock \href {https://www.aclweb.org/anthology/2020.emnlp-demos.6}
  {Transformers: State-of-the-art natural language processing}.
\newblock In \emph{Proceedings of the 2020 Conference on Empirical Methods in
  Natural Language Processing: System Demonstrations}, pages 38--45, Online.
  Association for Computational Linguistics.

\bibitem[{Xiong et~al.(2017)Xiong, Hoang, and Wang}]{xiong2017deeppath}
Wenhan Xiong, Thien Hoang, and William~Yang Wang. 2017.
\newblock Deeppath: A reinforcement learning method for knowledge graph
  reasoning.
\newblock In \emph{EMNLP}.

\bibitem[{Yang et~al.(2017)Yang, Yang, and Cohen}]{neural-lp}
Fan Yang, Zhilin Yang, and William~W Cohen. 2017.
\newblock Differentiable learning of logical rules for knowledge base
  reasoning.
\newblock In \emph{NIPS}.

\bibitem[{Yao et~al.(2019)Yao, Mao, and Luo}]{kgbert}
Liang Yao, Chengsheng Mao, and Yuan Luo. 2019.
\newblock Kg-bert: Bert for knowledge graph completion.
\newblock \emph{arXiv preprint arXiv:1909.03193}.

\end{thebibliography}
\bibliographystyle{acl_natbib}

\end{document}